\documentclass{article}
\usepackage{ijcai22}

\usepackage{times}

\usepackage{soul}
\usepackage{url}
\usepackage[hidelinks]{hyperref}
\usepackage[utf8]{inputenc}
\usepackage[small]{caption}
\usepackage{booktabs}
\usepackage[ruled]{algorithm}  
\usepackage[noend]{algpseudocode}
\usepackage{amsfonts}
\usepackage{amsmath}
\usepackage{amsthm}
\usepackage{bm}
\usepackage{float}
\usepackage{listings}
\usepackage{multirow}
\usepackage{graphicx}
\usepackage{xspace}
\usepackage{xcolor}

\makeatletter
\DeclareRobustCommand\onedot{\futurelet\@let@token\@onedot}
\def\@onedot{\ifx\@let@token.\else.\null\fi\xspace}

\def\eg{\emph{e.g\onedot}} 
\def\ie{\emph{i.e\onedot}} 
\def\cf{\emph{c.f}\onedot} 
 \def\vs{\emph{vs}\onedot}

\makeatother

\title{Residual Contrastive Learning for Image Reconstruction:\\
Learning Transferable Representations from Noisy Images}

\author{
Nanqing Dong$^1$\footnote{Contact Authors}\footnote{Part of this work done at Huawei Noah’s Ark Lab.}\and
Matteo Maggioni$^2$\and
Yongxin Yang$^2$\and
Eduardo Pérez-Pellitero$^2$\and\\
Ales Leonardis$^2$\and
Steven McDonagh$^2$$^*$
\affiliations
$^1$Department of Computer Science, University of Oxford\\
$^2$Huawei Noah’s Ark Lab
}

\begin{document}

\maketitle

\begin{abstract}
This paper is concerned with contrastive learning (CL) for low-level image restoration and enhancement tasks. We propose a new label-efficient learning paradigm based on residuals, \emph{residual contrastive learning} (RCL), and derive an unsupervised visual representation learning framework, suitable for low-level vision tasks with noisy inputs. While supervised image reconstruction aims to minimize residual terms directly, RCL alternatively builds a connection between residuals and CL by defining a novel instance discrimination pretext task, using residuals as the discriminative feature. Our formulation mitigates the severe task misalignment between instance discrimination pretext tasks and downstream image reconstruction tasks, present in existing CL frameworks. Experimentally, we find that RCL can learn robust and transferable representations that improve the performance of various downstream tasks, such as denoising and super resolution, in comparison with recent self-supervised methods designed specifically for noisy inputs. Additionally, our unsupervised pre-training can significantly reduce annotation costs whilst maintaining performance competitive with fully-supervised image reconstruction. 
\end{abstract}

\section{Introduction}
\label{sec:intro}
Fueled by the advances of self-supervised learning\footnote{We use the terms ``self-supervised learning'' and ``self-supervised representation learning'' interchangeably in this work.} (SSL), large-scale unsupervised pre-training followed by fine-tuning on small amounts of annotated data has become a popular label-efficient learning paradigm. A standard example involves firstly {unsupervised visual representation learning} (UVRL) using ImageNet~\cite{deng2009imagenet}. The learned representations can then be transferred to downstream tasks, reducing the number of labels required and yet achieving strong performance, competitive with supervised pre-training~\cite{he2020momentum}. Self-supervised strategies therefore have the potential to reduce labeling costs and this becomes especially pertinent for dense prediction tasks where human annotation is time-consuming and often very expensive. 

Further, an unsupervised pre-training method, capable of learning representations transferable across various downstream tasks, is desirable as this enables efficiency in terms of both computation and labeling costs. Without this property it quickly becomes impractical to label large datasets for multiple tasks of interest or alternatively design appropriate individual pretext tasks for each downstream application. 

\begin{figure}[t]
    \centering
    \includegraphics[width=0.6\columnwidth]{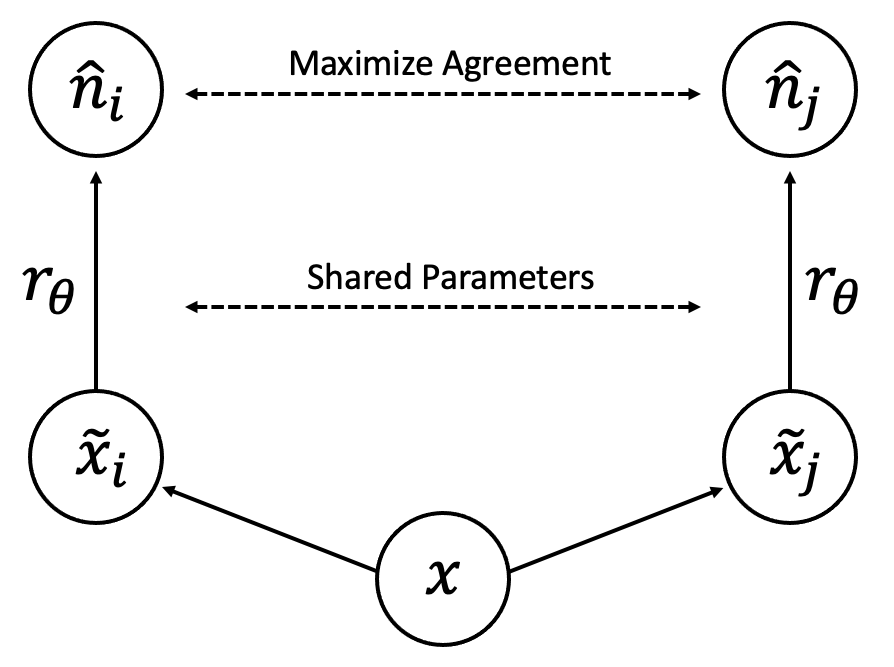}
    \caption{RCL for low-level visual representations with noisy inputs.
    $\bm{x}$ is a noisy image. $\tilde{\bm{x}}_i$ and $\tilde{\bm{x}}_j$ are two random crops from the same $\bm{x}$ (a positive pair). $r_\theta(\cdot)$ is a residual function, defined in Eq.~\ref{eq:residual}. $\hat{\bm{n}}_i$ and $\hat{\bm{n}}_j$ are two corresponding residual tensors.}
    \label{fig:framework}
\end{figure}

One such SSL strategy capable of learning transferable representations~\cite{wang2020understanding} is \emph{contrastive learning} (CL)~\cite{chen2020simple,he2020momentum,chuang2020debiased,li2021prototypical}. CL is based on an instance discrimination pretext task~\cite{wu2018unsupervised}, where the learning goal is to maximize the mutual information of two views of the same image, pulling two augmented image views together in the feature space, whilst pushing apart representations of different images~\cite{oord2018representation}. While CL has been shown to provide comparable performance with, and even improve upon, supervised pre-training with respect to various high-level downstream tasks~\cite{chen2020simple,he2020momentum}, contemporary strategies expose two limitations for low-level image restoration and enhancement tasks. 

Firstly, recent studies have shown that when the pretext and downstream tasks are not closely correlated, improving pretext task performance cannot guarantee downstream task improvement~\cite{ericsson2021well}. This phenomenon is known as \emph{task misalignment}~\cite{dong2021self}. However, existing CL frameworks are mainly designed for high-level semantic understanding tasks, leaving the potential of CL in conjunction with low-level vision domains currently underexplored. Empirically, we find that the task misalignment can severely impair the representation learning performance of existing CL frameworks for low-level downstream tasks. 

Secondly, images utilized for CL pre-training are typically assumed to be noise free, yet input to image enhancement and restoration tasks commonly contain additive noise. This is compounded by commonly adopted CL data augmentation policies~\cite{chen2020simple} that affect the data distribution and encourage the learning of invariances less relevant for image reconstruction tasks. 
Alternative SSL approaches have however been designed specifically for noisy images. Although such methods have achieved promising results for the denoising task specifically~\cite{batson2019noise2self}, we find that they are less well equipped to efficiently learn transferable representations when the downstream data distributions change (\ie~for additional low-level vision tasks).

Motivated by these considerations, our work aims to answer an under-explored question: \emph{how can CL be used to learn transferable representations for low-level vision tasks, from noisy images?} 

We start by recalling standard supervised learning (SL). Let ($x$, $y$) define an input and target image reconstruction respectively (\eg~a noisy and noise-free image pair in the denoising literature), the loss can then be formulated as $\|y - f_\theta(x)\|$, where $f_\theta(\cdot)$ is the model of interest with parameters $\theta$. This canonical use of paired data provides supervised models with a useful signal however obtaining ground truth data for real-world image enhancement and restoration tasks that necessitate dense prediction may require complex and often constraining procedures, \ie~$y$ is often unavailable due to annotation costs. Removing the requirement of a noise-free image $y$, and instead minimizing $\|x - f_\theta(x)\|$, can be seen to provide a trivial solution where $f_\theta(\cdot)$ is an identity mapping. Various SSL efforts therefore instead propose to minimize more useful objectives of the form $\|\tilde{x} - f_\theta(x)\|$, where $\tilde{x}$ constitutes \eg~a second noisy variant of $x$~\cite{lehtinen2018noise2noise,batson2019noise2self,ehret2019joint}. 
We observe that, without the norm operator, $x - f_\theta(x)$ can be regarded as a \emph{residual} term. In statistics and optimization, a \textit{residual} denotes the difference between observed and estimated values of interest. In the domain of deep learning residuals commonly take the form $r(x) = f_\theta(x) - x$, where $x$ is the input, $f_\theta(x)$ is the output, and $r(\cdot)$ is the residual function~\cite{he2016deep}. 

Following this formulation, we propose \emph{residual contrastive learning} (RCL), a residual-based SSL framework for noisy images (illustrated in Fig.~\ref{fig:framework}). We bridge a methodological gap between SSL on visual signals with additive noise and unsupervised residual learning via CL. We conjecture that residuals can be effectively used as a discriminative feature for CL based on the fact that additive image noise is signal-dependent~\cite{hasinoff2010noise}. We propose a \textit{residual contrastive loss}, which leverages the {earth mover's distance} (EMD) to measure the similarity between two residual tensors with the same shape (\cf~cosine similarity applied to two feature vectors~\cite{chen2020simple}). By leveraging signal-dependent noise as an appropriate discriminative feature based on the prior knowledge in image processing, in tandem with the representation learning ability of CL, RCL is expected to learn transferable representations amenable to downstream image reconstruction tasks.

Similar to previous CL studies that alternatively consider representation learning for high-level vision tasks~\cite{chen2020simple}, we adopt a \emph{proxy evaluation} protocol that uses the performance of {proxy} supervised downstream tasks to measure the quality of the representations learned during unsupervised pre-training. 
We establish a set of benchmark datasets and downstream tasks towards systematically evaluating RCL against both (i) recent CL methods that focus on dense prediction~\cite{pinheiro2020unsupervised,wang2021dense,xie2021propagate} and (ii) strong SSL methods designed specifically for noisy inputs~\cite{lehtinen2018noise2noise,batson2019noise2self,ehret2019joint}. We observe that representations learned by RCL consistently outperform the baselines and exhibit strong generalization ability in multiple downstream tasks; namely denoising, super resolution and demosaicing. Finally, we report that a learning paradigm involving pre-training on unlabeled data using RCL, followed by fine-tuning on small labeled data with SL, can achieve performance competitive with fully-supervised baselines.
In summary, our contributions are as follows: 
\\
\begin{enumerate}
\item~We provide the first formulation of an instance discrimination pretext task based on residuals.
\item~We propose RCL, a novel framework that can learn transferable representations from only noisy inputs. To the best of our knowledge, this constitutes the first study of CL on noisy images for low-level image reconstruction tasks.
\item~Our empirical results show that RCL learns robust representations from noisy images without paired ground truth, and unsupervised pre-training with RCL can significantly reduce the annotation cost in comparison with fully-supervised alternatives.
\end{enumerate}

\section{Related Work}
The recent renaissance of CL has been driven by the successes of UVRL on ImageNet~\cite{chen2020simple,he2020momentum}. In the case of image recognition, the objective of both the instance discrimination pretext task and corresponding downstream task are highly correlated and intuitively, so is resulting performance. However, for tasks such as object detection and others involving dense prediction, correlation still exists yet is found to be weaker than in the case of recognition~\cite{ericsson2021well}. To mitigate such task misalignment, several state-of-the-art (SOTA) CL frameworks, designed for dense prediction tasks, have been proposed, \eg~VADeR~\cite{pinheiro2020unsupervised}, DenseCL~\cite{wang2021dense}, and {PixContrast}~\cite{xie2021propagate}. These approaches perform pixel-wise CL and train encoder-decoder networks that directly enable dense prediction. However, in contrast to RCL, these SOTA approaches are designed and evaluated for \emph{semantic} understanding tasks, \ie~task misalignment still exists when the downstream task is related to image reconstruction. A further significant difference between this work and those highlighted is that we purposefully abstain from relying on data augmentation. We find that augmentation can alter the original data noise distribution and potentially provides a signal that leads to learning invariances, undesirable for our target low-level tasks.

There have additionally been recent applications of CL to \emph{specific} low-level vision tasks, \eg~dehazing~\cite{wu2021contrastive} and super resolution~\cite{wang2021towards}. In these studies, CL is considered as a regularization technique to provide an end-to-end solution for the specific task. In contrast, our proposed approach attempts to provide a step towards a \emph{universal} UVRL framework for unprocessed images with natural noise, instead of for task-dependent applications. Thus, RCL could be used as a pre-training step for these downstream tasks.  

A recent SSL study that, similar to our work, makes use of an EMD metric is \emph{self}-EMD~\cite{liu2020self}. In contrast to our proposed approach, their method formulates object detection as an optimal transport problem whereas we directly measure the similarity between two distributions without requiring an iterative procedure, commonly induced by optimal transport problems that make use of the Sinkhorn-Knopp algorithm.

\section{Residual Contrastive Learning}
\label{sec:rcl}

\subsection{Preliminary}
\label{sec:rcl:prelim}
A widely adopted contrastive loss, InfoNCE~\cite{oord2018representation}, is formulated as:
\begin{equation}
\mathcal{L}_{\text{NCE}} = - \log \frac{\exp(\mathrm{sim}(z_q, z_0) / \tau)}{\sum_{i=0}^{N} \exp(\mathrm{sim}(z_q, z_i) / \tau)}
\label{eq:nce}
\end{equation}
where $z$ denotes the feature vector extracted from an image patch of interest, $\tau$ is a temperature parameter, and $\mathrm{sim}(\cdot,\cdot)$ is the cosine similarity function.
Firstly image patches are encoded into feature vectors via an encoder and then $\mathrm{sim}(\cdot,\cdot)$ can be used to measure the similarity between these representations. $(z_q, z_0)$ is a positive pair such that two augmented views are taken from the same image; and $(z_q, z_{i > 0})$ is a negative pair, where two patches are taken from different images. 

\subsection{Problem Formulation}
\label{sec:rcl:problem}
We denote $\bm{x}$ as a noisy image signal with clean image signal $\bm{y}$ and additive noise $\bm{n}$. The relation\footnote{For simplicity, we assume $\bm{x}$ and $\bm{y}$ take a common image format, \eg~RGB or RAW. We discuss the special case that $\bm{x}$ and $\bm{y}$ have different image formats in Appendix~B.2.} between the tuple $(\bm{x}, \bm{y}, \bm{n})$ can be formed as $\bm{x} = \bm{y} + \bm{n}$. The noise element $\bm{n}$ follows an unknown signal-dependent distribution\footnote{See Appendix~A for discussion on signal-dependent noise.}.

For a low-level vision task under SL, a training set $\mathcal{S} = \{(\bm{x}_i, \bm{y}_i)\}_{i=1}^{N_\mathcal{S}}$ is given, with $N_\mathcal{S}$ training examples. Let $f_\theta$ denote a model of interest which takes $\bm{x}$ as input. The optimization goal is then to minimize $ \|f_\theta(\bm{x}) - \bm{y} \|_p$, for optimal model weights $\theta$ where $\|\cdot\|_p$ denotes the \emph{p}-norm. 

The problem setting of interest in this work is UVRL, which involves an unlabelled training set. We alternatively consider $\mathcal{S} = \{\bm{x}_i\}_{i=1}^{N_\mathcal{S}}$ and the goal is to learn representations (\ie~optimize model weights $\theta$) for downstream tasks with access to the noisy image signals only. 

\subsection{Residual-Based Instance Discrimination}
A key contribution of this study consists of the formulation of our \emph{residual}-based instance discrimination pretext task. We prime this by noting that supervised residual learning has led to success in many low-level vision tasks~\cite{zhang2017beyond,li2018multi}. The residual tensor for $\bm{x}$ is defined as 
\begin{equation}
    \hat{\bm{n}}(\bm{x}) = r_\theta(\bm{x}) = \bm{x} - f_\theta(\bm{x}).
    \label{eq:residual}
\end{equation}
We use the residual tensors as the discriminative input for CL. We are motivated by the empirical observation that on average, \emph{the noise distributions associated with two image crops, extracted from the same image, have detectably smaller divergence than noise distributions pertaining to crops extracted from different images}. This observation constitutes a natural extension of the signal-dependency assumption. Further, for natural images, the noise distributions of two crops originating from the same instance may also possess high correlation due to potential self-similarities~\cite{batson2019noise2self}, with similar structures appearing at different locations and scales in the same image.

\subsection{Residual Contrastive Loss}
\label{sec:rcl:loss}
We now formally introduce the proposed \textit{residual contrastive loss}. 
Note, the $\mathrm{sim}(\cdot,\cdot)$ function in Eq.~\ref{eq:nce} implicitly imposes two constraints: (i)~the input $z$ is required to take the form of a normalized vector; and (ii)~an element-wise correspondence between two feature vectors is required in the feature space. To realize a residual contrastive loss suitable for dense prediction tasks associated with low-level vision, we relax these constraints by replacing the cosine similarity $\mathrm{sim}(\cdot,\cdot)$ with a negative distance function. The original contrastive loss (Eq.~\ref{eq:nce}) can then be reformulated as 
\begin{equation}
\mathcal{L}_{\text{contrast}} = - \log \frac{\exp(-\mathrm{d}(\hat{\bm{n}}(\bm{x}_q), \hat{\bm{n}}(\bm{x}_0)) / \tau )}{\sum_{i=0}^{N} \exp( - \mathrm{d}(\hat{\bm{n}}(\bm{x}_q), \hat{\bm{n}}(\bm{x}_i)) / \tau)}
\label{eq:contrast}
\end{equation}
where $\tau$ is a temperature parameter~\cite{chen2020simple} and $\mathrm{d}(\cdot, \cdot)$ is a non-negative statistical metric measuring the divergence between two probability distributions, such that larger metric values indicate larger divergence.

\noindent \noindent\textbf{Distance Function}
We further note that, unlike cosine similarity, $\mathrm{d}(\cdot,\cdot)$ should not assume a pair-wise relationship between two samples, as the noise distribution is independent of the pixel location. Valid distance measures $\mathrm{d}(\cdot, \cdot)$ should also possess desirable properties such as ease of computation and differentiability, towards enabling efficient end-to-end training. Common information theoretic measures that require density estimation (\eg~\textit{Kullback–Leibler} divergence) do not meet the above requirements. In this work, we choose the \textit{earth mover's distance} (EMD).\footnote{We discuss alternative feasible options for $\mathrm{d}(\cdot,\cdot)$ in Appendix~B.1. We select EMD based on empirical results.}
Let $(\hat{\bm{n}}(\bm{x}_p), \hat{\bm{n}}(\bm{x}_q))$ be two residual tensors, we then have
\begin{equation}
\begin{split}
    &\text{EMD}(\hat{\bm{n}}(\bm{x})_p, \hat{\bm{n}}(\bm{x})_q) =\\ 
    &\inf_{\gamma \in \Pi(P_{p}, P_{q})}   \mathbb{E}_{(\hat{\bm{n}}(\bm{x})_p, \hat{\bm{n}}(\bm{x})_q) \sim \gamma}[\| \hat{\bm{n}}(\bm{x})_p -  \hat{\bm{n}}(\bm{x})_q \|],
\end{split}
\end{equation}
where $\hat{\bm{n}}(\bm{x})_p \sim P_p$, $\hat{\bm{n}}(\bm{x})_q \sim P_q$, and $\Pi(\cdot, \cdot)$ denotes the joint distribution. 

\noindent \noindent\textbf{Training}
For computational simplicity, we define a positive pair $(\hat{\bm{n}}(\bm{x}_q), \hat{\bm{n}}(\bm{x}_0))$ as two overlapping image patches $(\bm{x}_q, \bm{x}_0)$ cropped from the same instance and a negative pair $(\hat{\bm{n}}(\bm{x}_q), \hat{\bm{n}}(\bm{x}_{i>1}))$ as two image patches $(\bm{x}_q, \bm{x}_{i>1})$ cropped from two different instances. The batch-wise training details of the residual contrastive loss are illustrated in Algorithm~\ref{algo:1}.

\begin{algorithm}[t]
    \begin{algorithmic}[1]
    \State{\small Sample a batch of $N{+}1$ images.} \Comment{\small Sample $N{+}1$ positive pairs}
    \State{\small Sample two positive patches for each image.} 
    \State{\small Generate $\hat{\bm{n}}(\bm{x})$ for each of $2N{+}2$ patches.} \Comment{\small Eq.~\ref{eq:residual}}
    \For{\small $j = 1, 2, \ldots, N{+}1$} 
        \State{\small Take the $j^{\mathrm{th}}$ pair as the positive pair $(\hat{\bm{n}}(\bm{x}_q), \hat{\bm{n}}(\bm{x}_0))$.}
        \State{\small Take the $2^{\mathrm{nd}}$ patch of each of the other $N$ pairs as $\hat{\bm{n}}(\bm{x}_{i>1})$.}
        \State{\small Compute $\mathcal{L}_{\text{contrast}}$ for the $j^{\mathrm{th}}$ positive pair.} \Comment{\small Eq.~\ref{eq:contrast}}
    \EndFor
    \State{\small Sum up $\mathcal{L}_{\text{contrast}}$ for a batch of $N{+}1$ images as the batch-wise \textit{residual contrastive loss}.}
    \end{algorithmic}
    \caption{\small Batch-wise training of \textit{residual contrastive loss}}
    \label{algo:1}
\end{algorithm}

\subsection{Optimization}
\label{sec:rcl:opt}
While Eq.~\ref{eq:contrast} enables UVRL, this gives rise to a further question: as $f_\theta$ could represent an arbitrary function that satisfies Eq.~\ref{eq:residual}, the representations learned by RCL may not be meaningful for the downstream tasks of interest. CL works well for high-level visual representations because the pretext tasks and downstream tasks both involve discrimination of visual objects. Similarly, we require to build such a connection between RCL and low-level vision tasks.

The performance of low-level image reconstruction tasks is known to be sensitive to pixel-level intensities. This offers a simple solution: inclusion of the term \mbox{$\|\bm{x} - f_\theta(\bm{x})\|$} as a regularizer. Note that minimizing $\|\bm{x} - f_\theta(\bm{x})\|$ alone (\ie~without Eq.~\ref{eq:contrast}) could lead to the trivial solution of an identity mapping. This issue can be mitigated through the introduction of non-linearities to both terms. Inspired by this strategy, we leverage the basic concept of the \textit{perceptual loss}~\cite{johnson2016perceptual} and define a \emph{consistency} loss term as
\begin{equation}
    \mathcal{L}_{\text{consistency}} =\|\phi(g_e(\bm{x})) - \phi(g_e(f_\theta(\bm{x})))\|_2^2,
    \label{eq:perceptual}
\end{equation}
where $\phi(\cdot)$ represents the features extracted from a pre-trained encoder $g_e$. Note, $g_e$ could either be pre-trained in a self-supervised fashion using the unlabeled noisy inputs~\cite{he2020momentum} or acquired from existing pre-trained weights (\eg~from ImageNet~\cite{deng2009imagenet}). We utilize the assumption that the noisy input image and the reconstructed output image should convey similar semantic information, \ie~the noise should not drastically change the semantic content of the image. The final training objective is then the sum of the two introduced losses:
\begin{equation}
    \mathcal{L}_{\text{total}} = \alpha \mathcal{L}_{\text{contrast}} + \mathcal{L}_{\text{consistency}}, 
    \label{eq:total}
\end{equation}
where $\alpha$ is a weighting parameter chosen empirically.

\section{Experimental Setup}
\label{sec:exp:setup}
We introduce here experimental protocols, datasets and implementation details. We refer to Appendix~C for additional details on noise simulation, benchmark setup, and evaluations.

\noindent\textbf{Simulation} 
We aim to evaluate the generalization ability of the learned representations across different downstream tasks. However we note that real-world multi-task datasets, pertaining exclusively to low-level vision tasks, are currently scarce in the literature. Thus, to empirically validate the idea of CL with residuals, we firstly establish a set of benchmark datasets based on synthetic signal-dependent noise. To simulate such signal-dependent noise, we generate synthetic heteroscedastic Gaussian noise based on a noise level function\footnote{See Appendix A for the description of NLF.} (NLF) model ( $\bm{n} \sim \mathcal{N}(0, \lambda_{\mathrm{shot}} \bm{x} + \lambda_{\mathrm{read}})$). 
We use different ($\lambda_{\mathrm{shot}}$, $\lambda_{\mathrm{read}}$) to model different cameras and acquisition settings. The parameters ($\lambda_{\mathrm{shot}}$, $\lambda_{\mathrm{read}}$) are randomly sampled to ensure the overall noise variance level $\sigma^2$ of each image falls in a reasonable range for the data used in our experiments and we set $\sigma \in [0, 20]$ following~\cite{gharbi2016deep}. We therefore consider each image to have a noise distribution with approximately \textit{unique} parameters. From the perspective on the dataset $\mathcal{S}$, there is therefore an approximate one-to-one mapping between ($\lambda_{\mathrm{shot}}$, $\lambda_{\mathrm{read}}$) and each image. This simulation model is utilized to evaluate the robustness of SSL methods.

\noindent\textbf{Benchmark Datasets}
In order to simulate large-scale unlabeled training data with signal-dependent noise, we consider three large-scale public datasets, namely, the MIT Demosaicing dataset~\cite{gharbi2016deep} (MIT), the Stanford Taskonomy dataset~\cite{zamir2018taskonomy} (Stanford), and the PASCAL VOC dataset~\cite{everingham2010pascal} (VOC). We generate noisy images by adding synthetic noise. The datasets are split into training and test sets.

\noindent\textbf{Proxy Evaluation}
\label{sec:exp:proxy}
CL aims to learn strong representations for downstream tasks, \ie~pre-training of $f_\theta$ instead of solving each problem directly.  In this work, we therefore test the generalization ability of the learned representations~\cite{zhang2016colorful}.
Following previous studies on CL for high-level vision tasks~\cite{he2020momentum,chen2020simple,chuang2020debiased}, we adopt a \textit{proxy evaluation protocol}. Concretely, we fine-tune the learned representations on downstream tasks with a small amount of annotated data, under SL. We report the performance of the downstream tasks as the \textit{proxy performance} for SSL. In this way, we can systematically evaluate the generalization and transferablility of representations learned under different SSL frameworks. Following the \textit{linear classification protocol}~\cite{he2020momentum}, first the weights of a network $f_\theta$ are pre-trained using an unlabeled training set, and then all weights except those in the last layer are frozen. The pre-trained last layer is then replaced with a randomly initiated task-dependent layer for the downstream task. The new last layer is then fine-tuned with the labeled training set and evaluated on the task test set. Note, under proxy evaluation, the representations of the intermediate layers are fixed. The reported numerical results are used to indirectly reflect the quality of fixed representations, thus this is a \textit{proxy} evaluation. We highlight that this evaluation differs from common low-level vision task evaluation protocols, where end-to-end solutions are directly compared (without fine-tuning). 

\noindent\textbf{Evaluation Metrics}
We consider two common image reconstruction metrics for the proxy evaluation, \emph{peak signal-to-noise ratio} (PSNR) and \emph{structure similarity index measure} (SSIM). We repeat experiments over five trials and report mean results. We denote the performance of supervised pre-training as an \textit{Oracle}.

\noindent\textbf{Implementation} 
Theoretically, $f_\theta$ may constitute any model capable of performing dense prediction tasks. In the following section, we will show that the representations learned by $f_\theta$ can be successfully applied to various downstream image reconstruction tasks: denoising, demosaicing and super resolution. Following \cite{zamir2018taskonomy}, we utilize a generic network backbone; U-Net~\cite{ronneberger2015u} to instantiate $f_\theta$. We additionally use a ResNet50~\cite{he2016deep}, pre-trained on ImageNet~\cite{deng2009imagenet} for the fixed feature extractor $g_e$. To instantiate Eq.~\ref{eq:contrast}, we follow~\cite{chen2020simple} in defining temperature $\tau$ values and use a batch size of $64$. We use a weighting parameter $\alpha{=}10^{-3}$ in the unsupervised pre-training phase and an $L_1$ loss for the supervised fine-tuning in the evaluation phase. We use an Adam~\cite{kingma2015adam} optimizer with $\beta_1{=}0.9$, $\beta_2{=}0.999$, and $\epsilon{=}10^{-7}$, and a fixed learning rate $10^{-3}$. The minimal image crop size is $128{\times}128$. All models are implemented in PyTorch on a NVIDIA Tesla V100 GPU.

\section{Experiments}
\label{sec:exp:rgb}
\noindent\textbf{Baselines}
To validate the empirical considerations presented in Sec.~\ref{sec:intro}, we select two sets of baselines SSL methods.  We include three SOTA CL frameworks for high-level vision tasks to validate our hypothesis that there exists a task misalignment between semantic understanding tasks and image restoration tasks. We consider VADeR~\cite{pinheiro2020unsupervised}, DenseCL~\cite{wang2021dense}, and {PixContrast}~\cite{xie2021propagate}. These three baselines apply CL at a pixel-level in the feature space, thus can train an encoder-decoder network directly for dense prediction tasks. However, in contrast to RCL, these methods are designed for semantic understanding tasks, \ie~task misalignment still exists. We use a consistent U-Net backbone for all three methods, where the number of output channels of the last layer is set to three (for RGB images). The pre-training and fine-tuning procedures follow Sec.~\ref{sec:exp:proxy}, inline with our RCL. For a fair comparison, we also report the performance of three methods trained with additional $\mathcal{L}_{\text{consistency}}$ (Eq.~\ref{eq:perceptual}), denoted as VADeR+, DenseCL+, and {PixContrast}+.
We also consider two seminal SSL baselines that are designed for noisy images, namely \textit{noise2noise} (N2N)~\cite{lehtinen2018noise2noise} and \textit{noise2self} (N2S)~\cite{batson2019noise2self}, for UVRL on image reconstruction tasks. For N2N, we generate paired noisy RGB images with the same random parameters ($\lambda_{\mathrm{shot}}$, $\lambda_{\mathrm{read}}$). Note that N2N and N2S both utilize a formulation of $\|\tilde{\bm{x}} - f_\theta(\bm{x})\|$, where $\tilde{\bm{x}}$ is a noisy observation of $\bm{x}$.

\begin{table}
  \centering
  \caption{Proxy evaluation of representation learning using denoising as the downstream task. }
  \label{tab:rgb_denoising}
  {\scriptsize
  \begin{tabular}{lrrrrrr}
    \multirow{2}{*}{Method} & \multicolumn{2}{c}{MIT} & \multicolumn{2}{c}{Stanford} & \multicolumn{2}{c}{VOC} \\
    & PSNR & SSIM & PSNR & SSIM & PSNR & SSIM\\ \hline
    VADeR & 14.63 & 0.1088 & 17.54 & 0.1601 & 16.33 & 0.1573\\
    DenseCL & 13.78 & 0.0910 & 16.46 & 0.1527 & 15.33 & 0.1373\\
    \emph{PixContrast} & 14.77 & 0.1101 & 17.61 & 0.1610 & 16.42 & 0.1585\\ \hline
    VADeR+ & 19.63 & 0.4183 & 21.58 & 0.4961 & 20.75 & 0.4737\\
    DenseCL+ & 18.87 & 0.3998 & 20.58 & 0.4705 & 20.21 & 0.4647\\
    \emph{PixContrast}+ & 19.87 & 0.4121 & 21.41 & 0.4899 & 20.82 & 0.4858\\
    N2N & 28.66 & 0.8614 & 34.14 & 0.8699 & 30.91 & 0.8272\\
    N2S & 28.16 & 0.8373 & 34.04 & 0.8640 & 30.71 & 0.8256\\ \hline
    RCL-BD & 28.83 & 0.8871 & 34.75 & 0.8618 & 31.29 & 0.8274 \\
    RCL-MMD & 28.68 & 0.8864 & 34.87 & 0.8687 & 31.53 & 0.8316\\ 
    RCL-EMD & \noindent\textbf{29.54} & \noindent\textbf{0.8908} & \noindent\textbf{35.43} & \noindent\textbf{0.8783} & \noindent\textbf{31.39} & \noindent\textbf{0.8330}\\\hline
    \textit{Oracle} & 31.26 & 0.9187 & 38.25 & 0.9422 & 33.65 & 0.9038\\\hline
   \end{tabular}
    }
\end{table}

\noindent\textbf{Denoising}
We instantiate denoising as the first downstream task and report representation learning results in Table~\ref{tab:rgb_denoising}. All three CL frameworks, designed for high-level tasks, produce results much weaker than SSL methods designed for specific low-level vision tasks, with or without $\mathcal{L}_{\text{consistency}}$. We emphasize that this is due to the highlighted severe task misalignment issue. 
Thus these methods are omitted in the following discussion. Note, during proxy evaluation, the pre-training set and testing set do not overlap. RCL shows competitive representation learning performance in comparison with N2N and N2S, which are reported to achieve reasonable performance in blind denoising tasks. In addition to EMD, we consider two alternative distance functions \textit{Bhattacharyya distance} (BD) and \textit{maximum mean discrepancy} (MMD). EMD showed more robust performance than BD and MMD in denoising and two other downstream tasks (below), we thus select EMD as our default metric for remaining experiments.\footnote{Additional details found in Appendix~B.1 and D.1.}

\noindent\textbf{Super Resolution}
We further explore the generalization ability of our learned representations to low-level vision tasks that are markedly distinct from denoising. Super resolution (SR) constitutes one such task. As each image in the Stanford dataset has two resolutions ($512{\times}512$ and $1024{\times}1024$), we define the higher resolution image as the upsampled ground truth in order to provide a simple proof of concept SR task. Following the proxy evaluation protocols introduced previously, results are presented in Table~\ref{tab:rgb_sr} (left). 
\begin{figure}[t]
   \centering{
   \captionsetup{type=table}
\caption{Proxy evaluation of representation learning using SR and JDenSR as the downstream task on the Stanford dataset.}
        \label{tab:rgb_sr}
        {\small
        \begin{tabular}{lcccc}
        \multirow{2}{*}{Method} & \multicolumn{2}{c}{SR} & \multicolumn{2}{c}{JDenSR}\\ 
        & PSNR & SSIM & PSNR & SSIM\\\hline
        N2N &  31.61 & 0.8730 & 27.90 & 0.7860 \\
        N2S &  31.11 & 0.8699 & 27.80 & 0.7831\\ \hline
        RCL-BD & 38.89 & 0.9654 & 32.10 & 0.8046 \\
        RCL-MMD & 38.31 & 0.9634 & 31.95 & 0.8068 \\
        RCL-EMD & \textbf{39.01} & \textbf{0.9658} & \textbf{32.63} & \textbf{0.8214} \\ \hline
        SL (Den) & 34.18 & 0.9118 & 32.89 & 0.8353\\
        \textit{Oracle} & 38.93 & 0.9603 & 35.98 & 0.9175\\\hline
        \end{tabular}
        }
        }
\end{figure}
Here we can observe that RCL outperforms N2N and N2S by a large margin. By comparing Table~\ref{tab:rgb_denoising} with Table~\ref{tab:rgb_sr}, we find the performance gap between RCL and N2N (N2S) becomes larger, \ie~N2N and N2S tend to learn less meaningful representations for disparate downstream tasks where the gap between them and respective pretext tasks grow, in the investigated setting. This phenomenon has also been discussed in~\cite{zhang2016colorful}, where a task-dependent colorization-based SSL shows limited performance in image classification. As a comparison, RCL can result in representations that exhibit stronger generalisation ability.  

\noindent\textbf{Joint Denoising and Super Resolution}
As a natural extension to independent denoising and SR tasks, we consider joint denoising and super resolution (JDenSR) as the downstream task, which has two sub-tasks and can further demonstrate the versatility of the learned representations. The results are presented in Table~\ref{tab:rgb_sr} (right). Again, RCL outperforms the baseline SSL frameworks by a large margin. We note that the objective for image reconstruction tasks is typically to estimate (minimize) residuals. This is a similar setup to denoising but with the only difference that the residual might have a different distribution. We hypothesize that RCL is more robust to this change of distribution. 

\noindent\textbf{Transferability: Supervised Pre-Training \vs Unsupervised Pre-Training}
In addition to unsupervised pre-training, we report the performance of supervised pre-training by denoising in the ``SL (Den)'' row  of Table~\ref{tab:rgb_sr}. We learn representations by applying SL to the denoising task, defined in Table~\ref{tab:rgb_denoising}. We then fine-tune to the alternative downstream tasks in a fashion identical to the considered SSL methods. We note that interestingly, RCL is able to outperform ``SL (Den)'' for the SR task and also RCL(-EMD) achieves higher performance than the \textit{Oracle} in Table~\ref{tab:rgb_sr}. This unintuitive phenomenon, that unsupervised pre-training can improve performance over supervised pre-training, has been recently corroborated in CL studies that consider high-level vision tasks~\cite{wang2020understanding}. While supervised pre-training tends to learn task-dependent representations, the representations learned by CL are more informative. In Table~\ref{tab:rgb_sr}, the ``SL (Den)'' row, pertaining to JDenSR results, exhibits strong performance, and a marginal advantage over RCL, which may be explained by the fact that JDenSR can be considered closely related to a pure denoising task.

\noindent\textbf{RCL \vs SL}
To further quantify the performance and labelling-cost trade-off, we perform a sensitivity study. We train a U-Net in a supervised fashion (SL) for the denoising task using VOC data and compare this with RCL(-EMD) under various magnitudes of available training labels. We re-use the \textit{same} random seeds for both methods.
In the first row of Table~\ref{tab:rgb_pretrain}, we report the performance of RCL by directly applying the representations pre-trained on the unlabelled training set, on the test set. In the remaining Table~\ref{tab:rgb_pretrain} rows, it can be observed that the performance gain obtained by pre-training with RCL grows larger as SL suffers more from label scarcity. 

\begin{table}[t]
        \centering
        \caption{Standard SL (left) and RCL pre-training with SL fine-tuning (right), evaluated with denoising on the VOC dataset. \# Labels denotes the number of labeled data available for SL.}
        \label{tab:rgb_pretrain}
        {\small
        \begin{tabular}{lrrrr}
        \multirow{2}{*}{\# Labels} & \multicolumn{2}{c}{SL} & \multicolumn{2}{c}{RCL + SL}\\
        & PSNR & SSIM & PSNR & SSIM\\ \hline
        0 & - & - & 22.62 & 0.7989\\\hline
        $10$ & 20.74 & 0.7299 & 28.20 & 0.8834 \\
        $10^2$ & 27.19 & 0.8734 & 30.24 & 0.9028 \\
        $10^3$ & 31.31 & 0.9184 & 32.09 & 0.9280 \\
        $10^4$ & 33.41 & 0.9437 & 33.85 & 0.9514 \\\hline
        \end{tabular}
        }
\end{table}

\noindent\textbf{Label-Efficient Learning}
Our sensitivity study affords us some initial evidence towards answering the questions: \emph{can RCL help SL?} and, if so, \emph{when can RCL help?} We fine-tuned the U-Net, pre-trained by RCL(-EMD) on the entire training set, with additional paired RGB training images, as above. 
Pre-training with RCL consistently improves the performance of standard SL. In cases where labelled data are rare, expensive to collect or curate, such pre-training may be able to offer significant improvement (\eg~$+7.46dB$ with only ten labels in Table~\ref{tab:rgb_pretrain}). We observe that fine-tuning with $500$ and $4000$ labels achieves similar performance to supervised training with $10^3$ and $10^4$ labels, which are around $50\%$ and $60\%$ reduction in terms of annotation cost, respectively. We can also observe that improvement margins diminish as the number of available labels grow significantly (\eg~$+0.44dB$ with $10,000$ labels).

\noindent\textbf{Effect of Residual Contrastive Loss}
Following the same training procedure of N2N and N2S, we minimize $\mathcal{L}_{\text{consistency}}$ (Eq.~\ref{eq:perceptual}) alone to validate the contribution of the residual contrastive loss. The result of minimizing $\mathcal{L}_{\text{consistency}}$ alone is lower than N2N and N2S but higher than VADeR+, DenseCL+, and {PixContrast}+, which are negatively impacted by the task misalignment effect. It is worth mentioning that including the residual contrastive loss in the training can not significantly improve the model robustness or generalization ability for different downstream tasks, as shown in Table~\ref{tab:rgb_sr}.

\begin{figure}[t]
        \centering
        \captionsetup{type=figure}
        \includegraphics[width=0.7\columnwidth]{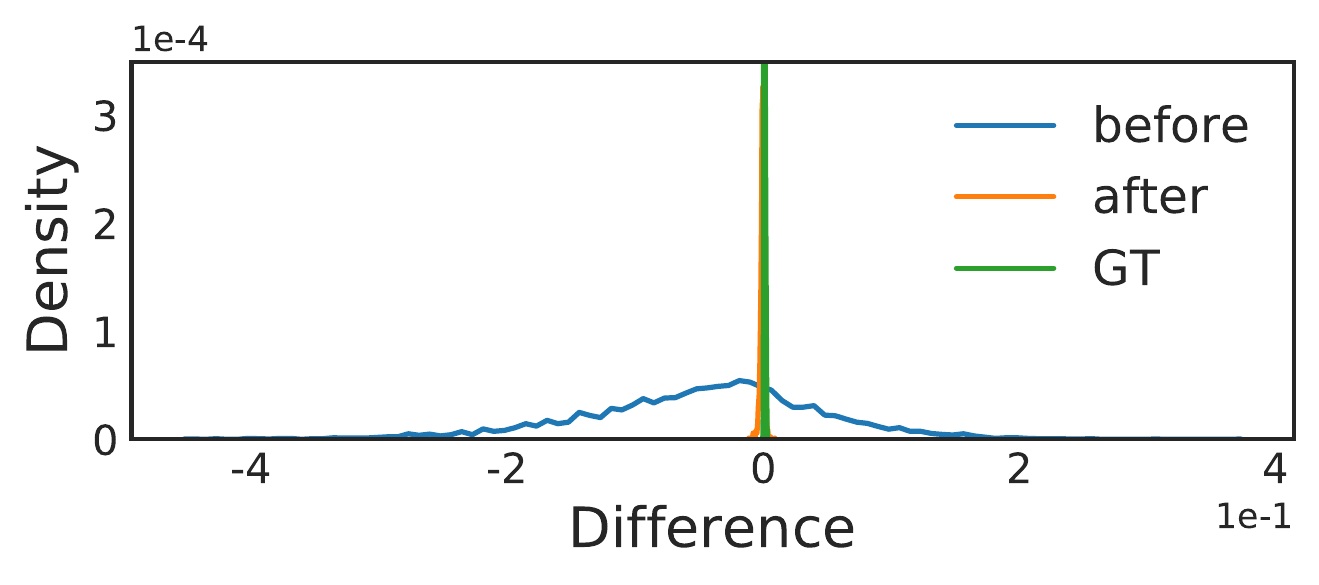}
        \caption{Comparison of residual contrastive loss (illustrated with EMD) before and after training. The density plot depicts the pair-wise differences of EMD between negative pairs minus EMD between positive pairs.}
        \label{fig:density}
\end{figure}

\noindent\textbf{Learning from Residuals}
It is important to validate that RCL indeed learns from the residuals in the proposed formulation. To illustrate the learning outcome directly, we extract the residual tensors by using a U-Net trained on the MIT dataset with RCL-EMD. Given an anchor image, we calculate the pair-wise difference for EMD between a negative pair and EMD between a positive pair. Given the same network, we record the differences before the training starts (\ie~the weights are randomly initialized) and after the loss converges. The density plot of the differences is shown in Fig.~\ref{fig:density}. RCL contracts the predicted distribution closer to the true underlying distribution, where we use the sampled noise as the residual. We also find that employing large $\alpha$ values in Eq.~\ref{eq:total} degrade the performance. 
We conjecture that this is because low-level vision tasks are sensitive to pixel-level perturbation. To provide an example: a minor change in predicted pixel intensity can change the reconstructed pixel color but an analogous change in predicted pixel probability may not meaningfully change \eg~a segmentation result. 
RCL with large $\alpha$ can still learn representations, however these may not be appropriate for the downstream tasks, discussed in Sec.~\ref{sec:rcl:opt}.

\noindent\textbf{Additional Results} We include additional ablation studies, experiments concerning three further downstream tasks related to RAW image data, and initial evidence that RCL can be efficiently utilized to reduce annotation costs in a real-world denoising task in Appendix~D.

\noindent\textbf{Limitations}
The empirical results in Sec.~\ref{sec:exp:rgb} are based on simulation. It is interesting to evaluate the proposed framework on the real-world multi-task datasets in the future.

\section{Conclusion}
\label{sec:conclusion}
We present a principled unsupervised strategy which can learn transferable representations from images with additive noise for different image reconstruction tasks. 
To the best of our knowledge, we are the first to unify CL and residual learning by formulating a residual-based instance discrimination pretext task.
The empirical studies validate the robustness and generalization of the representations learned by RCL, and further pose a new generic and label-efficient learning direction for low-level vision tasks. 

\bibliographystyle{named}
\bibliography{refs}

\clearpage
\appendix
\section{Signal-Dependent Noise Model}
In the domains of computer vision and image processing, given an observed image $\bm{x}$ and its underlying noise-free image $\bm{y}$ serving as ground truth, we have 
\begin{equation}
    \bm{x} = \bm{y} + \bm{n}, \bm{n} \sim P(\bm{n})
    \label{eq:noise}
\end{equation}
where $\bm{n}$ constitutes additive noise following a probability distribution $P(\bm{n})$, that is introduced during the image acquisition process due to hardware limitations. The real $P(\bm{n})$ is typically unknown in practical problems and has been modelled with varying levels of complexity~\cite{foi2008practical,wei2020physics}. A common simplifying assumption involves statistically modelling $P(\bm{n})$ as a homoscedastic Gaussian distribution. This simplification makes a strong assumption that the noise is independent of the underlying signal.
However, many previous studies~\cite{healey1994radiometric,gow2007comprehensive,liu2008automatic,foi2008practical,hasinoff2010noise,makitalo2012optimal} have rigorously shown that image noise is in fact signal-dependent; $\bm{n} \sim P(\bm{n} | \bm{y})$, \ie~that the probability distribution of $\bm{n}$ is conditioned on $\bm{y}$. Statistical signal-dependent noise modeling has thus also been previously explored and a common modelling choice is the zero-mean heteroscedastic Gaussian model~\cite{mohsen1975noise,liu2014practical}, known as the noise level function (NLF) in the domain of computational photography, 
\begin{equation}
    n_i \sim \mathcal{N}(0, \lambda_{\mathrm{shot}} x_i + \lambda_{\mathrm{read}}),
    \label{eq:nlf}
\end{equation}
where $n_i$ is the noise at pixel $y_i$ of the clean image $\bm{y}$, and  $\lambda_{\mathrm{shot}}$ and $\lambda_{\mathrm{read}}$ are parameters controlling the variance of the Gaussian model.

\section{Residual Contrastive Learning}
\subsection{Alternative Distance Functions}
\subsubsection{Bhattacharyya Distance}
The \textit{Bhattacharyya distance} (BD) has a closed-form expression in the case that the two distributions of interest are Gaussian. The sample distribution of residual tensors can be approximated by a Gaussian distribution when the crop size is large enough.
Formally, the BD between two Gaussian distributions $\hat{\bm{n}}(\bm{x}_p){\sim} \mathcal{N}(\mu_p, \sigma^2_p)$ and $\hat{\bm{n}}(\bm{x}_q){\sim} \mathcal{N}(\mu_q, \sigma^2_q)$ can be estimated as 

\begin{equation}
\begin{split}
    \mathrm{BD}(\hat{\bm{n}}(\bm{x}_p), \hat{\bm{n}}(\bm{x}_q)) &= \frac{1}{4} \ln (\frac{1}{4} (\frac{\sigma^2_p}{\sigma^2_q} + \frac{\sigma^2_q}{\sigma^2_p} + 2))\\ &+ \frac{1}{4} (\frac{(\mu_p - \mu_q)^2}{\sigma^2_p + \sigma^2_q}), 
\end{split}
    \label{eq:bd}
\end{equation}
where $\mu$ and $\sigma^2$ are the sample mean and the sample variance. 

\subsubsection{Maximum Mean Discrepancy}
The \textit{maximum mean discrepancy} (MMD) is a non-parametric method based on the \textit{kernel embedding of distributions} where a probability distribution is represented as an element of a {reproducing kernel Hilbert space}~\cite{gretton2012kernel}. Given a domain $\Omega$, let an arbitrary function $h:\Omega \to \mathbb {R}$ belong to a class of functions $\mathcal{H}$, MMD is mathematically defined as
\begin{equation}
    \mathrm{MMD}(\mathcal{H}, P_p, P_q) = \underset{h \in \mathcal{H}}{\mathrm{sup}}(\mathbb{E}_p[h(\hat{\bm{n}}(\bm{x}_p))] - \mathbb{E}_q[h(\hat{\bm{n}}(\bm{x}_q))]).
\end{equation}
Similar to EMD, MMD does not make any assumption over the distribution in question and has a robust empirical estimation based on Gaussian kernels~\cite{pan2010domain}.

\subsection{Ill-Defined Residual}
\label{app:jdd}
In Sec.~4 of our main manuscript, the difference $\bm{x} - f_\theta(\bm{x})$ implicitly assumes that $\bm{x}$ and $f_\theta(\bm{x})$ (or $\bm{y}$) have the same shape, or more specifically, reside in the same color space. A common scenario in low-level vision is that the input and output signals reside in different color spaces. One such example low-level image reconstruction task of fundamental importance is joint demosaicing and denoising (JDD)~\cite{hirakawa2006jdd,khashabi2014joint,gharbi2016deep}, which transforms noisy RAW images to clean RGB images.
In such cases $\bm{x}$ now defines an image in RAW format, and $f_\theta$ learns a mapping for our considered example task, JDD.
Given a RAW input $\bm{x}$ with shape $H{\times}W{\times} 1$\footnote{Following~\cite{gharbi2016deep}, a practical option is to first rearrange $\bm{x}$ into shape $H/2{\times}W/2{\times}4$ as a pre-processing step, where each channel contains one of the four colors in the CFA array.}, the output $f_\theta(\bm{x})$ will produce an RGB image with shape $H{\times}W{\times}3$. Such a formulation unfortunately results in Eq.~2 becoming ill-defined. One simple solution involves redefinition of the residual term in the RAW domain. We can thus redefine Eq.~2 as $\hat{\bm{n}}(\bm{x}) = \bm{x} - \mathrm{mosaic}(f_\theta(\bm{x}))$, where $\mathrm{mosaic}$ is a standard image mosaic operation that transforms an image in RGB format to RAW format. In lieu of applying $g_e$, pre-trained directly in the RAW domain, we adopt a simple consistency loss formulation for JDD~\cite{ehret2019joint}:
\begin{equation}
 \mathcal{L}_{\text{consistency}} =\|\bm{x} - \mathrm{mosaic}(\mathcal{T}(f_\theta(\bm{x})))\|   
 \label{eq:jdd}
\end{equation}
where $\mathcal{T}$ is a linear interpolation operation~\cite{keys1981cubic}. Note, due to the mosaic operation lacking an inverse, Eq.~\ref{eq:jdd} does not offer a trivial solution.

\section{Experimental Setup}
\subsection{Noise Simulation}
\label{app:exp:noise}
Typically, the shot and read noise parameters ($\lambda_{\mathrm{shot}}$, $\lambda_{\mathrm{read}}$) for any given camera sensor approximately follow a log-linear relationship in the parameter space~\cite{foi2008practical,brooks2019unprocessing}. The specific values of these parameters for a given camera and acquisition settings can be automatically obtained through automatic calibration procedures~\cite{foi2008practical} and can be found in the metadata pertaining to the RAW images for most cameras. In~\cite{brooks2019unprocessing}, the authors use a number of ($\lambda_{\mathrm{shot}}$, $\lambda_{\mathrm{read}}$) samples to define the log-linear distributions of the noise in the Darmstadt Noise Dataset (DND)~\cite{plotz2017benchmarking}. Such distribution is further perturbed along the log-line by adding a small Gaussian perturbation which models the specific variability of DND. For example, the simulation model for RAW data in \cite{brooks2019unprocessing} is 
\begin{align}
    \log(\lambda_{\mathrm{shot}}) &\sim \mathcal{U}(\log(0.001), \log(0.12)),\\
    \log(\lambda_{\mathrm{read}}) &\sim \mathcal{N}(2.18 \log(\lambda_{\mathrm{shot}}) + 1.2, 0.26).
\end{align}

In our case, we need to go one step further as we would like to fully randomize the relation of the noise parameters in the ($\lambda_{\mathrm{shot}}$, $\lambda_{\mathrm{read}}$) domain so as to simulate noise from different cameras and different acquisition settings.
We randomly sample ($\lambda_{\mathrm{shot}}$, $\lambda_{\mathrm{read}}$) in the parameter space accordingly. 

Given $\mathcal{U}(\sigma_{\mathrm{min}}, \sigma_{\mathrm{max}})$, where $\sigma_{\mathrm{min}}^2$ and $\sigma_{\mathrm{max}}^2$ are the minimum possible variance and maximum possible variance, the simulation model is
\begin{align}
    a = \sqrt{2 (\frac{\sigma_{\mathrm{min}}}{2})^2}, \ & b = \sqrt{2 (\frac{\sigma_{\mathrm{max}}}{2})^2}, \\
    \sqrt{\lambda_{\mathrm{shot}}} \sim \mathcal{U}(a, b), \ &
    \sqrt{\lambda_{\mathrm{read}}} \sim \mathcal{U}(a, b).
\end{align}
The noise simulation model presented above will sample ($\lambda_{\mathrm{shot}}$, $\lambda_{\mathrm{read}}$) for Eq.~\ref{eq:nlf}. The minimum possible variance $\sigma_{\mathrm{min}}^2$ and the maximum possible variance $\sigma_{\mathrm{max}}^2$ enable us to fully control the strength of the noise applied to the input. Note that our parameterization, in contrast to~\cite{brooks2019unprocessing}, will not constrain ($\lambda_{\mathrm{shot}}$, $\lambda_{\mathrm{read}}$) to follow a pre-defined log-linear relationship, but instead can sample the parameters at arbitrary positions in the domain hence naturally enabling us to simulate noise from different camera sensors.

\subsection{Simulation Datasets}
The three datasets considered in the main manuscript contain RGB images with varying semantic information and image quality. MIT is a benchmark demosaicing dataset, containing images that are collected from the web with large variety in terms of object categories and scenes. Stanford is a benchmark multi-task learning (MTL) dataset, which consists of various indoor scenes. VOC is semantic segmentation dataset of $20$ semantic categories. The images of MIT and VOC have various resolutions, while the images of Stanford have two fixed resolutions, $512{\times}512$ and $1024{\times}1024$.  
We select two subsets of MIT and Stanford, and VOC to validate our ideas. The MIT subset contains $11000$ images\footnote{\url{https://groups.csail.mit.edu/graphics/demosaicnet/dataset.html}}, 
divided into $10000$ images in the training set and $1000$ images in the test set. The Stanford subset contains $9464$ images\footnote{\url{https://github.com/alexsax/taskonomy-sample-model-1}}, and we use $8464$ images for training and $1000$ images for testing. VOC contains 17125 images\footnote{\url{http://host.robots.ox.ac.uk/pascal/VOC/}}, including $16125$ for training and $1000$ for testing. The minimal crop size is $128{\times}128$ for the Stanford dataset and only $64{\times}64$ for the MIT and VOC datasets, limited by the original image size.

\subsubsection{Simulated RAW Datasets}
For JDD, following previous work~\cite{heide2014flexisp,gharbi2016deep,ehret2019joint}, each RGB image is first mosaiced to form a Bayer pattern (\eg~\texttt{RGGB} in this work) RAW image and then synthetic noise is added. 

\subsection{Evaluation Metrics}
\subsubsection{Peak Signal-to-Noise Ratio}
PSNR is defined as 
\begin{equation}
    \mathrm{PSNR} = 10 \log_{10} \frac{255^2}{\frac{1}{H \times W \times C} \sum_i^H \sum_j^W \sum_k^C (x - y)^2},
\end{equation}
where $H$, $W$, $C$ are the height, width, and number of channels for paired images $x$ and $y$. We use the default implementation of \texttt{scikit-image} package\footnote{\url{https://scikit-image.org/docs/dev/api/skimage.metrics.html}}.

\subsubsection{Structure Similarity Index Measure}
SSIM is defined as 
\begin{equation}
    \mathrm{SSIM} = l(x - y)^{\alpha} c(x - y)^{\beta} s(x - y)^{\gamma},
\end{equation}
where
\begin{equation}
\begin{split}
    \alpha > 0,& \ \beta > 0, \ \gamma > 0, \nonumber\\
    l(x - y) &= \frac{2 \mu_x \mu_y + c_1}{\mu_x^2 + \mu_y^2 + c_1}, \nonumber\\
    c(x - y) &= \frac{\sigma_{xy} + c_2}{\sigma_x^2 + \sigma_y^2 + c_2}, \nonumber\\
    s(x - y) &= \frac{\sigma_{xy} + c_3}{\sigma_x \sigma_y + c_3}.
\end{split}
\end{equation}
Here, $\mu_x$ and $\mu_y$ are the mean of $x$ and $y$, $\sigma_x$ and  $\sigma_y$ are the standard deviation of $x$ and $y$, $\sigma_{xy}$ is the covariance between $x$ and $y$, and $c_1$, $c_2$, and $c_3$ are constants.
We use the default implementation of \texttt{scikit-image} package.

\section{Additional Experiments}
\subsection{Ablation Studies}
\subsubsection{Distance Metric}
Among the three investigated distance metrics, EMD empirically shows robust performance in comparison to BD and MMD. RCL with EMD consistently outperforms RCL with BD and MMD (Table 1 and Table 2). Our primary focus has been to present the overall RCL framework and theoretical analysis on the choice of the best distance metric is considered beyond the scope of the current work. Based on the empirical results, EMD may be preferred in practical denoising applications. However, we would note that RCL with BD and MMD also show competitive performance with respect to baselines, which gives support to the notion of RCL efficiency as a learning framework. 

\subsubsection{Crop Size \vs Batch Size} Under the constraint of limited computational resource, there will be a trade-off between crop size and batch size. Intuitively, a larger sample size should improve the estimation of the statistical distance. We do observe a performance drop of $2.06~dB$ in PSNR when RCL-EMD is trained with a crop size reduced from $128{\times}128$ to $64{\times}64$ on the Stanford dataset. However, increasing the batch size under crop size $64{\times}64$ also leads to a decrease in performance (\eg~$-0.98~dB$ when batch size is increased from 64 to 128). This contradicts theoretical findings that imply a larger batch size is always preferred in CL~\cite{chuang2020debiased}. As \cite{chuang2020debiased} focus on image classification problems where images are distinct, we conjecture that this phenomenon may be caused by the fact that different residual tensor pairs may have similar scalar values by coincidence (\cf~a feature vector cosine similarity does not have this concern). A large batch size may also increase the chance of such coincidence.

\subsection{Experiments on RAW Data}
\label{app:exp:jdd}
\subsubsection{Baselines} We additionally evaluate RCL with RAW image data, following a similar protocol as the RGB data. There is limited related work in the literature. We use \textit{mosaic2mosaic} (M2M)~\cite{ehret2019joint}, a state-of-the-art SSL JDD framework, as the SSL baseline. 
In this section, we use RCL to denote RCL-EMD, which was reported to have the most robust performance in Sec.~5 of the main paper. 

\begin{table}[t]
  \centering
  \caption{Proxy evaluation of representation learning using JDD, JDemSR, and JDDSR as the downstream tasks on Stanford data.}
  \label{tab:app:raw}
  {\scriptsize
  \begin{tabular}{lrrrrrr}
    \multirow{2}{*}{Method} & \multicolumn{2}{c}{JDD} & \multicolumn{2}{c}{JDemSR} & \multicolumn{2}{c}{JDDSR} \\
        & PSNR & SSIM & PSNR & SSIM & PSNR & SSIM \\\hline
    M2M & 26.62 & 0.7463 & 27.21 & 0.8219 & 26.07 & 0.7292\\
    RCL & \noindent\textbf{27.37} & \noindent\textbf{0.7514} & 28.00 & 0.8160 & 26.87 & 0.7331\\\hline
    SL (JDD) & - & - & \noindent\textbf{32.55} & \noindent\textbf{0.8727} & \noindent\textbf{30.70} & \noindent\textbf{0.8323} \\
    \textit{Oracle} & 33.60 & 0.8979 & 36.63 & 0.9431 & 31.56 & 0.8490 \\\hline
   \end{tabular}
    }
\end{table}

\subsubsection{Downstream Tasks} Following Sec.~5, the representations are evaluated with three related downstream tasks that concern RAW data, which consist of joint demosaicing and denoising (JDD), joint demosaicing and super resolution (JDemSR), and joint demosaicing, denoising and super resolution (JDDSR). The results are evaluated on Stanford data and are presented in Table~\ref{tab:app:raw}. While RCL outperforms M2M overall, the performance gap between RCL and the baseline becomes smaller while the gap between RCL and SL becomes larger, in comparison with Sec.~5 of the main manuscript. We believe this to be partially caused by the challenging ill-posed inverse nature of the demosaicing reconstruction task. The redefined residuals between $\bm{x}$ and $f_\theta(\bm{x})$ in Appendix~\ref{app:jdd} no longer have a pixel-level correspondence, which may hamper RCL from learning transferable representations.

\subsection{Label-Efficient Learning on Real Noisy Data}
\label{sec:exp:real}
\subsubsection{Dataset}
The Smartphone Image Denoising Dataset (SIDD)~\cite{abdelhamed2018high} is a camera image dataset with real additive noise\footnote{\url{https://www.eecs.yorku.ca/~kamel/sidd/index.php}}.
SIDD contains 160 noisy and clean image pairs, with ten individual scenes repeatedly captured using five different smartphone cameras. We use the SIDD-Medium Dataset, which is a subset of SIDD. Compared with the simulated signal-dependent data, SIDD has less variation in terms of ($\lambda_{\mathrm{shot}}$, $\lambda_{\mathrm{read}}$).

\subsubsection{Preparation}
To illustrate the practical value of RCL, we apply it to denoising of real noisy camera data. In this experiment, we aim to show that RCL can be efficiently utilized to reduce the annotation cost using the learning paradigm of unsupervised pre-training followed by supervised fine-tuning. We use the RGB data from the Smartphone Image Denoising Dataset (SIDD)~\cite{abdelhamed2018high} where $80\%$ of the data is considered as the training set and the final $20\%$ is reserved for testing. The data is rearranged into patches of size $128{\times}128$. In order to show that the formulation of RCL generalizes well, we let $f_\theta$ be a DnCNN~\cite{zhang2017beyond}, a state-of-the-art denoiser. Note however that our learning paradigm can be considered agnostic to specific denoising architectures. While common approaches predict the reconstructed pixels directly (\eg~U-Net in Sec.~5), DnCNN alternatively predicts the residuals\footnote{Under this formulation, representations will have a lower generalization ability for other downstream tasks, \cf~UNet in Sec.~5}. In contrast to~Eq.~2, the residual term now takes the form of $\bm{x} - (\bm{x} - f_\theta(\bm{x}))$, such that Eq.~3 and Eq.~5 still hold. We use a batch size of 64 and an Adam optimizer with a constant learning rate $10^{-3}$. We use a somewhat rudimentary training procedure as the main purpose of the experiment is to validate the label-efficient learning paradigm, rather than competing quantitatively with the state-of-the-art. 

\begin{table}[t]
    \centering
    \caption{Label-efficient learning on real noisy data, evaluated with denoising on the SIDD dataset. SL (left) denotes SL with trained with the full labeled training set. RCL + SL (right) denotes RCL pre-training with full unlabeled training set with SL fine-tuning with only $30\%$ of labeled data.}
    \label{tab:app:real}
    {\footnotesize
    \begin{tabular}{rrrr}
        \multicolumn{2}{c}{SL} & \multicolumn{2}{c}{RCL + SL}\\
        PSNR & SSIM & PSNR & SSIM \\ \hline
        37.21 & 0.936 & 37.64 & 0.944
    \end{tabular}
    }
\end{table}

\subsubsection{Results}
Following the same procedure used to generate the results found in Table~3 of the main manuscript, the goal is to illustrate that unsupervised pre-training with RCL can improve the performance of SL with the same amount of annotated data; or in other words, unsupervised pre-training with RCL can reduce the annotation cost in comparison with fully supervised alternatives. 
As the baseline, we train a DnCNN with the full training set in a supervised fashion. For label-efficient learning, we first train another DnCNN with the full training set in an unsupervised fashion via RCL. Then, the DnCNN is fine-tuned with only $30\%$ of the labeled training set.
The results are presented in Table~\ref{tab:app:real}. We find that, with only $30\%$ of the labeled data, label-efficient learning can outperform standard SL using the full training set and efficiently reduce the annotation cost. The results in Table~3 in the main paper and Table~\ref{tab:app:real} suggest that the representations obtained via RCL pre-training can slightly improve the performance while using only a limited fraction of available labels. We believe that this is due to the fact that, while standard SL may result in overfitting, the representations learned by RCL generalize well on the unseen test data.

\end{document}